\documentclass[letterpaper, 10 pt, journal, twoside]{IEEEtran}
\IEEEoverridecommandlockouts                                
\newcounter{myromancnt}

\UseRawInputEncoding

\usepackage{graphicx, cite, bm}
\usepackage{amsmath}
\usepackage{xcolor}
\usepackage{amsfonts}
\usepackage{multirow}
\usepackage{mathtools}
\usepackage{color}
\usepackage{cite}
\usepackage{amssymb,amsfonts}
\usepackage{graphicx}
\usepackage{textcomp}
\usepackage{xcolor}                                   
\usepackage{kotex}      
\usepackage{algpseudocode}
\usepackage{algorithm}
\usepackage{amsmath,bm}
\usepackage{tabularx}
\usepackage[caption=false,font=footnotesize]{subfig}
\DeclareMathOperator{\diag}{diag}
\usepackage{dblfloatfix}
\usepackage[hidelinks]{hyperref}
\usepackage{siunitx}
\usepackage{textcase}
\usepackage{tikz}
\usepackage{textcomp}
\usepackage{hyperref}
\usepackage{lipsum}

\newcommand\copyrighttext{%
  \footnotesize \textcopyright 2025 IEEE.  Personal use of this material is permitted.  Permission from IEEE must be obtained for all other uses, in any current or future media, including reprinting/republishing this material for advertising or promotional purposes, creating new collective works, for resale or redistribution to servers or lists, or reuse of any copyrighted component of this work in other works.
  DOI: \href{https://ieeexplore.ieee.org/document/10884016}{10.1109/LRA.2025.3541428}}
\newcommand\copyrightnotice{%
\begin{tikzpicture}[remember picture,overlay]
\node[anchor=south,yshift=5pt] at (current page.south) {\fbox{\parbox{\dimexpr\textwidth-\fboxsep-\fboxrule\relax}{\copyrighttext}}};
\end{tikzpicture}%
}

\begin{document}
\bstctlcite{IEEEexample:BSTcontrol}










\title{ Online Friction Coefficient Identification for Legged Robots on Slippery Terrain Using Smoothed Contact Gradients}

\author{Hajun Kim$^{1}$, Dongyun Kang$^{1}$, Min-Gyu Kim$^{1}$, Gijeong Kim$^{1}$ and Hae-Won Park$^{1}$, \textit{Member, IEEE}

\thanks{Manuscript received: June, 11, 2024; Revised November, 20, 2024; Accepted January, 26, 2025.}
\thanks{This paper was recommended for publication by Editor Abderrahmane Kheddar upon evaluation of the Associate Editor and Reviewers' comments.
This work was supported in part by Korea Evaluation Institute of Industrial Technology (KEIT) funded by the Korea Government (MOTIE) under Grant No.20018216, Development of mobile intelligence SW for autonomous navigation of legged robots in dynamic and atypical environments for real application.)} 
\thanks{
$^{1}$Authors are with the Humanoid Robot Research Center, School of Mechanical, Aerospace \& Systems Engineering, Department of Mechanical Engineering, Korea Advanced Institute of Science and Technology (KAIST), Yuseong-gu, 34141 Daejeon, Republic of Korea. {\tt\small haewonpark@kaist.ac.kr}}
\thanks{Digital Object Identifier (DOI): see top of this page.}

}

\markboth{IEEE Robotics and Automation Letters. Preprint Version. Accepted JANUARY, 2025}
{Kim \MakeLowercase{\textit{et al.}}: Online Friction Coefficient Identification for Legged Robots on Slippery Terrains Using Smoothed Contact Gradients}

\maketitle

\copyrightnotice


\begin{abstract}
  This paper proposes an online friction coefficient identification framework for legged robots on slippery terrain. The approach formulates the optimization problem to minimize the sum of residuals between actual and predicted states parameterized by the friction coefficient in rigid body contact dynamics. Notably, the proposed framework leverages the analytic smoothed gradient of contact impulses, obtained by smoothing the complementarity condition of Coulomb friction, to solve the issue of non-informative gradients induced from the nonsmooth contact dynamics. Moreover, we introduce the rejection method to filter out data with high normal contact velocity following contact initiations during friction coefficient identification for legged robots. To validate the proposed framework, we conduct the experiments using a quadrupedal robot platform, KAIST HOUND, on slippery and nonslippery terrain. We observe that our framework achieves fast and consistent friction coefficient identification within various initial conditions.


\end{abstract}


\section{Introduction} \label{Sec:Introduction}
\IEEEPARstart{F}{or} legged robots navigating challenging terrain, contact modeling for considering the interaction between the robot and terrain is crucial. The modeling is particularly critical on slippery terrain, where the robots encounter nonlinear and hybrid dynamics due to foot slippage.
Recently, contact modelings using rigid body contact dynamics have gained attention in the field of legged robots~\cite{wensing2024opti,lidec2023contact,mujoco,raisim}. 

However, the friction coefficient, a critical parameter for Coulomb friction in contact dynamics that substantially influences the dynamics' propagation~\cite{acosta2022validating}, often has its estimated value different from the actual one based on the terrain. Consequently, contact dynamics modeling with an inaccurately estimated friction coefficient can diverge further from real-world dynamics~\cite{varin2020constrained}.
Although the friction coefficient critically affects the solution of contact dynamics, identifying this parameter poses challenges due to the nonlinear and nonsmooth nature of contact dynamics, particularly during slip events.

Over the years, researchers have actively explored the derivatives of nonsmooth dynamics~\cite{tolsma2002hidden,kong2024saltation}. More recently, for dynamics parameter identification, some studies~\cite{werling2021fast,NEURIPS2018_842424a1,lidec2022differentiable} have focused on differentiable physics simulators that offer the gradients with respect to dynamics parameters. The given gradient is then utilized for gradient-based strategies to handle the optimization problem of system identification. Especially, the authors of~\cite{lidec2022differentiable,jatavallabhula2021gradsim} demonstrated the friction coefficient identification using real collected data. Their approach focused on offline identification, verifying their frameworks on simple systems like a sliding box.

\begin{figure}
    \centering
    \subfloat[]{
        \includegraphics[width=0.55\columnwidth]{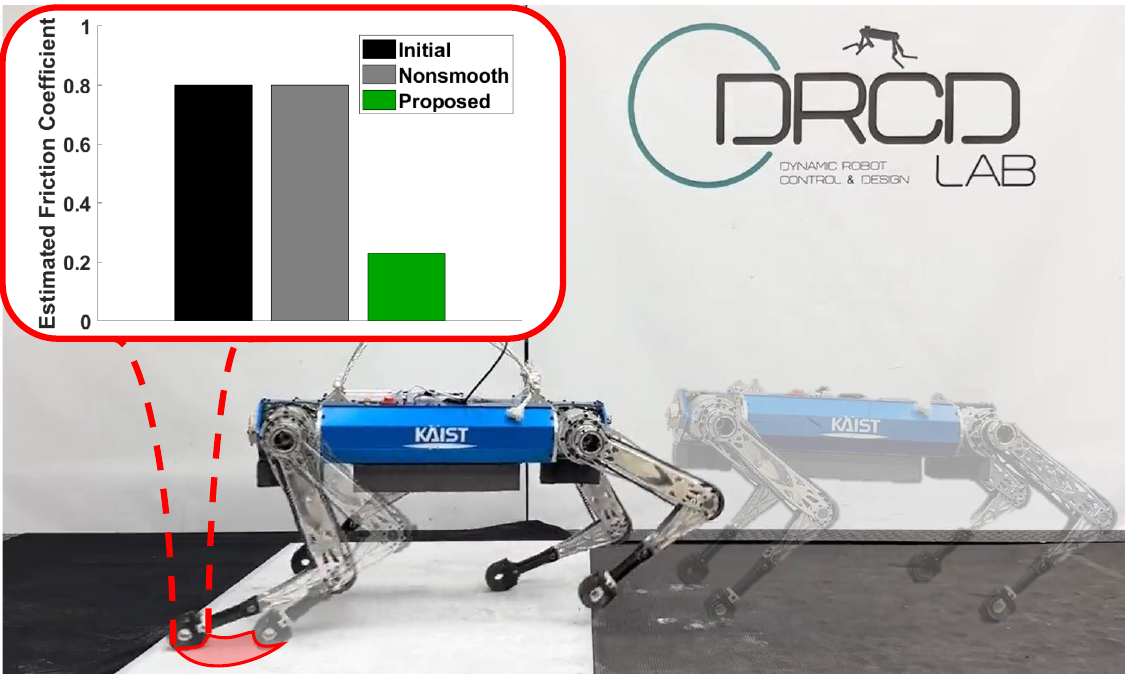}
        \label{fig:a}
        }
\hfill
    \subfloat[]{
        \includegraphics[width=0.39\columnwidth]{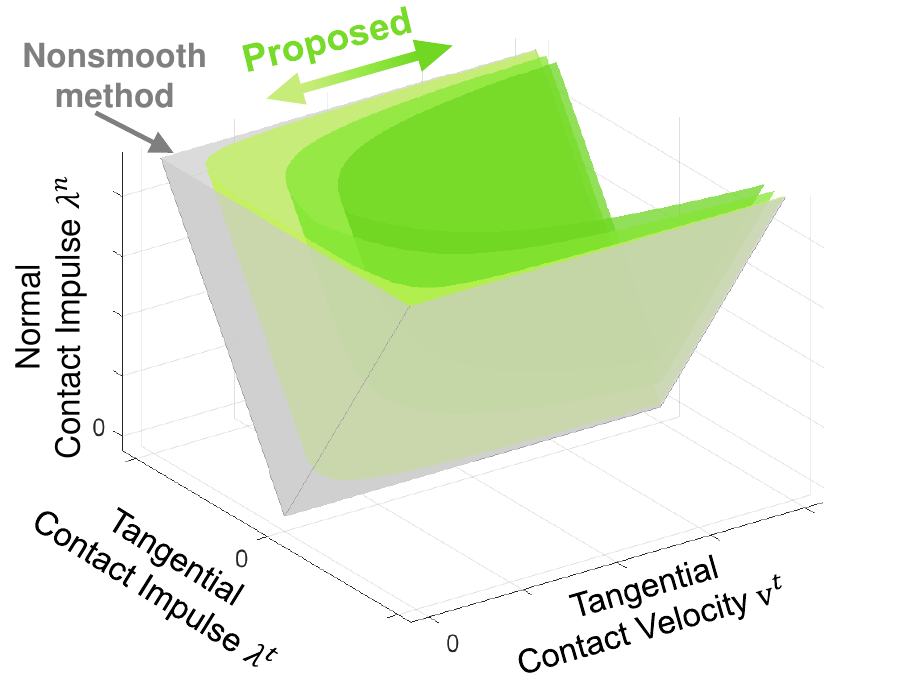}
        \label{fig:b}
        }
    \caption{We present an online friction coefficient identification framework using proprioceptive measurements for legged robots. \protect\subref{fig:a} With proposed smoothed gradients with respect to the friction coefficient, our framework can handle the issue of non-informative gradients caused by the nonsmooth contact dynamics in friction coefficient identification \protect\subref{fig:b} An illustration of constraint space for proposed gradients and nonsmooth gradients.}
    \label{fig:wrong_cof_in_contact_model}
\end{figure}

However, due to the nonsmooth constraints in contact dynamics, exact analytic gradients often become non-informative~\cite{le2024leveraging}. This issue can impede the exploration for
the lower loss solution in the optimization process due to the numerical challenges posed by the complementarity condition~\cite{werling2021fast}.
To tackle the issue, various studies have proposed smoothing techniques~\cite{zachary2024fast,kim2022contact,le2024leveraging,Pang2023TRO}. Recently, the author in~\cite{le2024leveraging} adopted the randomized smoothing used in reinforcement learning to mitigate the nonsmoothness issue in optimal control systems. In particular, the authors in~\cite{Pang2023TRO} showed that applying randomized smoothing methods to motion planning exhibited performance comparable to that of analytic smoothing. However, they also observed that those methods required longer computation times than the analytic smoothing method since they rely on the sampling methods.

For the legged robot's friction coefficient identification, the authors in~\cite{yu2017preparing} demonstrated the network-based friction coefficient identification framework in simulation. Furthermore, the study in~\cite{focchi2018slip} validated the friction coefficient estimation in simulation, using ground reaction forces estimated through the legged robot's joint torque measurements. The work in ~\cite{jenelten2019legged} proposed a slip estimation framework that considers kinematics during slip events but not friction-related dynamics. In~\cite{lee2020learning}, a learning-based decoder was used to restore friction coefficients in slippery terrains, while achieving a robust controller. This approach required extensive training time with simulation data, and did not provide gradient information for parameters during inference. Recently, the study~\cite{chen2022real} proposed an online system identification method that uses a confidence score-based update integrated with a model predictive controller. This study introduced the confidence score for various parameters, including the friction coefficient, to assess the confidence of data used for system identification. However, the study verified system identification for several parameters, except for the friction coefficient, using robot arms with unknown end-effector mass distribution.

In this study, we aim to develop an online friction coefficient estimation framework for legged robots using proprioceptive measurements. The main contributions of this paper can be summarized as follows:
\begin{itemize}
\item We propose analytic smoothed gradients of contact impulses with respect to the friction coefficient to tackle the lack of informative gradients. 
\item We introduce a rejection method that excludes data with high normal contact velocity while using the confidence score-based parameter updates of~\cite{chen2022real}.
\item Through our proposed methods, the friction coefficient identification for legged robots can achieve faster and more consistent performance than using randomized smoothing~\cite{Pang2023TRO,le2024leveraging} and the nonsmoothed approach~\cite{chen2022real}.
\item We validate the proposed framework with the KAIST HOUND quadrupedal robot hardware~\cite{shin2022hound}.
\end{itemize}


The remainder of this paper is organized as follows: Section \ref{Sec:Background} introduces the background of this study. Section \ref{Sec:Method} details the proposed methodologies, and Section \ref{Sec:Experiment} describes experimental results. Finally, Section \ref{Sec:Discussion} and \ref{Sec:Conclusion} presents the discussion and conclusion, respectively.


\section{Background} \label{Sec:Background}
\begin{figure}
    \centering
    \includegraphics[width=1.0\columnwidth]{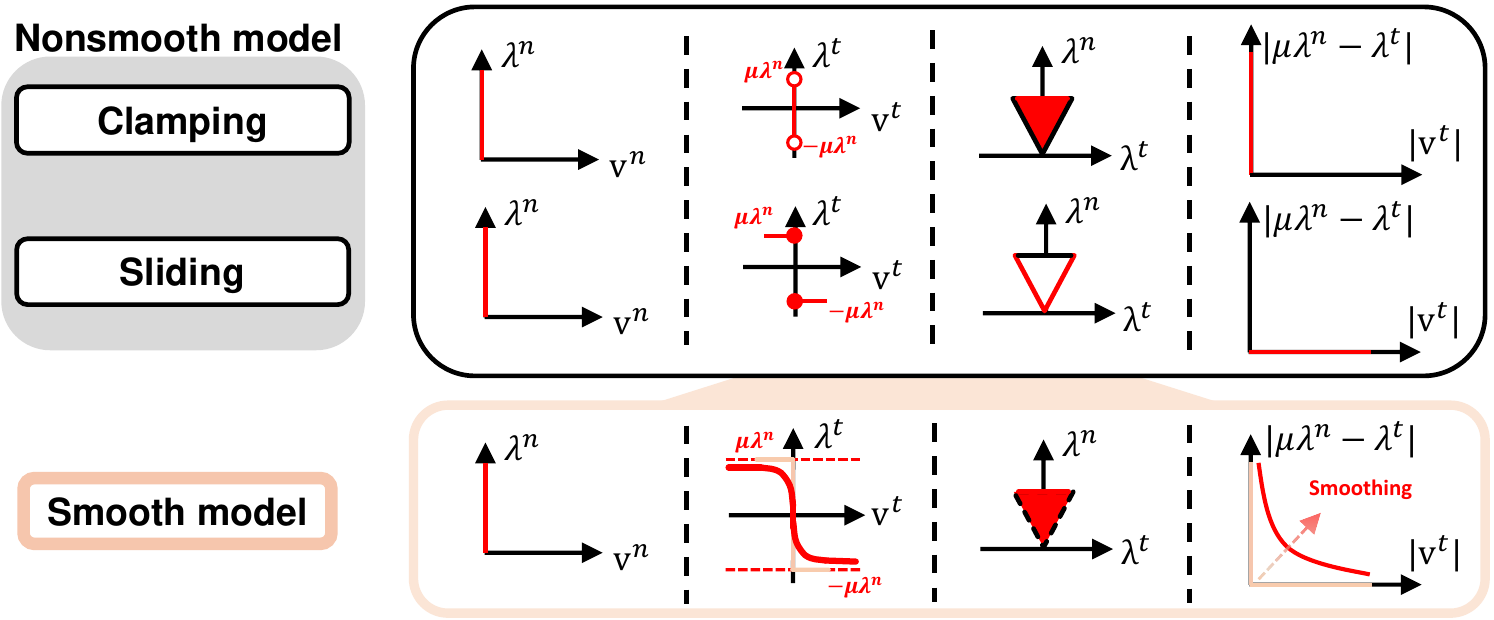}
    \caption{An illustration of contact states covered in rigid-body contact dynamics, excluding an opening contact, and our proposed smoothed conditions. The proposed smoothing is applied to the complementarity condition of Coulomb friction in contact states. In the smoothed conditions, the red line represents the smoothed constraint, while the brown line depicts the nonsmoothed constraint.}
    \label{figure:contact_dynamics}
\end{figure}
Our approach is based on the method of the previous works~\cite{raisim,werling2021fast,kim2022contact,chen2022real}. In this section, based on the studies, we introduce an optimization problem for system identification, contact dynamics, and analytic gradients of the contact impulse with respect to the friction coefficient. 

\subsection{Optimization Problem for System Identification}
Consider a discrete-time dynamic model as follows:
\begin{align}
\label{eq:dynamics}
    \mathbf{\hat{x}}_{i+1} & = f(\mathbf{x}_{i},\mathbf{u}_{i},\theta),
\end{align}
where $\mathbf{x}$ is the generalized states and $\mathbf{u}$ is the control input, $\theta$ is the parameter of dynamics, and $f$ is the propagation function of dynamics. 
With the history buffer of states, the optimization for system identification can be defined as follows\cite{chen2022real}:
\begin{equation}
\label{osi_opt_basic}
     \theta^{*}=\arg\min_{\theta} \frac{1}{2}\sum_{i=1}^{H-1} \left\| {\mathbf{\hat{x}}}_{i+1}-\mathbf{x}_{i+1}\right\|^2,
\end{equation}
where $\theta$ is the parameter to be identified and $H$ is the size of buffer. To obtain $\theta^{*}$, the gradient-based strategy can be adapted using a step size of $\alpha$, with $\Delta{\theta}=-\alpha\mathbf{G}$. The gradient of the loss consists of the product of residuals and derivatives with respect to the parameter: $\mathbf{G}=\sum_{i=1}^{H-1} (\frac{df(\mathbf{x}_i,\mathbf{u}_{i},\theta)}{d\theta})^{T}({\mathbf{\hat{x}}}_{i+1} - \mathbf{x}_{i+1})$. Note that if the dynamics is independent of the parameter, for example, $\frac{df(\mathbf{x}_i,\mathbf{u}_{i},\theta)}{d\theta}=0$, which represents a non-informative gradient, the parameter updates,  $\Delta{\theta}$, can go to zeros or become non-informative.

\subsection{Frictional Contact Dynamics}
Consider an articulated rigid body system in contact with its environment. The discrete-time dynamic model of the system is as follows:
\begin{align}
\label{eq:dynamics}
    \mathbf{q}_{i+1} & = \mathbf{q}_i + \mathbf{\boldsymbol{\upsilon}}_{i+1}\Delta t\nonumber,\\
\mathbf{\boldsymbol{\upsilon}}_{i+1} & = \mathbf{M}^{-1}((-\mathbf{h}+\mathbf{B}\boldsymbol{\tau}_i)\Delta t + \mathbf{M}\mathbf{\boldsymbol{\upsilon}}_i+\mathbf{J}^T\bm{\lambda}),
\end{align}
 where $\mathbf{q}\in\mathbb{R}^{n_q}$ is generalized coordinate, $\boldsymbol{\upsilon}\in\mathbb{R}^{n_{\upsilon}}$ is generalized velocity, $\boldsymbol{\tau}\in\mathbb{R}^{n_a}$ is the generalized torque, $\mathbf{M}\in\mathbb{R}^{n_{\upsilon}{\times}n_{\upsilon}}$ represents the joint space inertial matrix, ${\mathbf{h}} \in\mathbb{R}^{n_{\upsilon}}$ accounts for Coriolis, centrifugal and gravitational terms, ${\mathbf{J}}\in\mathbb{R}^{3n_{c}{\times}n_{\upsilon}}$ is the contact Jacobian, $\mathbf{B}$ is an input matrix, and $\Delta t$ is a time step. ${\bm{\lambda}}$ is the vector consisting of contact impulses $\bm{\lambda}_{k}$ at each contact point where ${k=1,\cdots,n_{c}}$.  
Each contact impulse $\bm{\lambda}_{k}$ consists of normal components $\lambda^{n}_{k}$ and tangential components $\lambda^{t}_{k}$. Similarly, each contact Jacobian and contact velocity can be distinguished into normal and tangential components.

The relation between contact velocity and contact impulse can be described as follows:
\begin{align}
\label{eq:contactvel}
    \mathbf{v}_{k,i+1}=\mathbf{\sigma}_k+\mathbf{A}_{k} \bm{\lambda}_{k},
\end{align}
where $\mathbf{\sigma}_k:=\mathbf{J}_k \mathbf{M}^{-1} ((-\mathbf{h}+\mathbf{B} \boldsymbol{\tau}_i)\Delta t +\mathbf{J}_{\tilde{k}}^T \bm{\lambda}_{\tilde{k}}+\mathbf{M} \boldsymbol{\upsilon}_i)$. 
$\mathbf{A}_{k}:=(\mathbf{J}_k \mathbf{M}^{-1} \mathbf{J}_k^T)^{-1}$ is the apparent inertia matrix at $k$-th contact point, $\tilde{k}$ denotes indices except for $k$, and $\textbf{v}_{k,i+1}$ is the contact velocity at the $k$-th contact point for the next time step.

The contact impulse $\bm\lambda_{k}$ and contact velocity $\textbf{v}_{k,i+1}$, governed by the following conditions and principle constraints—the Signorini condition: $0\leq\text{g}^{n}_{k,i+1}\perp{{\lambda}^{n}_{k}}\geq0$ where $\text{g}^{n}_{k,i+1}$ is a gap between $k$th contact bodies, Coulomb's friction cone constraint parameterized by the friction coefficient $\mu$: $\| \lambda^{t}_{k} \|_{2} \leq \|\mu\lambda^{n}_{k}\|_{2} $, and the maximum dissipation principle—allow for the classification of contact states as shown in Fig.~\ref{figure:contact_dynamics}. The Signorini condition for velocity level can be employed for indices of closed contacts: $0\leq\text{v}^{n}_{k,i+1}\perp{{\lambda}^{n}_{k}}\geq0$.
The Maximum Dissipation Principle states that contact forces are chosen to maximize the dissipation of kinetic
energy.
The contact impulse can be calculated by solving the following optimization problem~\cite{moreau1977application}:
\begin{align}
\label{eq:minvel}
    \min_{\bm{\lambda}_k}{\mathbf{v}_{k,i+1}}^T \mathbf{M}_{k}{\mathbf{v}_{k,i+1}}\\
    s.t. \quad \bm{\lambda}_k\in {\cal{S}}_{\mu}\nonumber,
\end{align}
where ${\cal{S}}_\mu$ is defined by the feasible set of elements satisfying the Signorini condition and Coulomb's friction cone constraint.

\renewcommand{\arraystretch}{1.2}
\begin{table}[t!]
\centering
\caption{Comparison between the proposed and the nonsmooth model for the gradient of contact impulses with respect to the friction coefficient.}
\resizebox{\columnwidth}{!}{%
\Large
\begin{tabular}{|c|c|c||cc|cc|}
\hline
\multirow{2}{*}{\textbf{Cases}}                                & \multirow{2}{*}{\textbf{\begin{tabular}[c]{@{}c@{}}Contact state \\ in dynamics\end{tabular}}} & \multirow{2}{*}{\textbf{\begin{tabular}[c]{@{}c@{}}Complementarity\\ Constraints\end{tabular}}} & \multicolumn{2}{c|}{\textbf{Nonsmooth model}}                                                                                                            & \multicolumn{2}{c|}{\textbf{Proposed}}        \\ \cline{4-7}      &                                                                                                      &                                                                                                 & \multicolumn{1}{c|}{\textbf{Gradients}}                                                                                    & \textbf{Updates}                            & \multicolumn{1}{c|}{\textbf{Gradients}}                                                                                            & \textbf{Updates}                            \\\hline \hline
\multirow{4}{*}{\begin{tabular}[c]{@{}c@{}}Actual Slipping\\ 
($\mu_\mathrm{true} \ll \hat{\mu}) $\end{tabular}} & \multirow{2}{*}{Clamping}                                                                            & $\|\textbf{v}^{t}\| = 0$                                                                        & \multicolumn{1}{c|}{\multirow{2}{*}{\textbf{\begin{tabular}[c]{@{}c@{}}$ {\frac{\partial\boldsymbol{\mathbf{\lambda}}}{\partial\mu}}$ = $\mathbf{0}$ \\ (non-informative) \\  \end{tabular}}}} & \multirow{2}{*}{$\|{\Delta\hat{\mu}\| = 0}$}            & \multicolumn{1}{c|}{\multirow{2}{*}{\textbf{\begin{tabular}[c]{@{}c@{}}$ {\frac{\partial\boldsymbol{\mathbf{\lambda}}}{\partial\mu}} \neq \mathbf{0}$ \\ (Informative) \\ \end{tabular}}}} & \multirow{2}{*}{$\|{\Delta\hat{\mu}\|>0}$ } \\ \cline{3-3}     &                                                                                                      & \textbf{$\hat{\mu}\lambda^{n} \textgreater \|\lambda^{t}\|$}                                                              & \multicolumn{1}{c|}{}                                                                                            &                                    & \multicolumn{1}{c|}{}                                                                                                    &                                    \\ \cline{2-7}           & \multirow{2}{*}{Sliding}                                                                             & $\|\textbf{v}^{t}\| > 0 $                                                            & \multicolumn{1}{c|}{\multirow{2}{*}{$\frac{{\partial\boldsymbol{\mathbf{\lambda}}}}{{\partial\mu}} \neq \mathbf{0}$}}                                                          & \multirow{2}{*}{$\|\Delta\hat{\mu}\|>0$} & \multicolumn{1}{c|}{\multirow{2}{*}{${\frac{\partial\boldsymbol{\mathbf{\lambda}}}{\partial\mu}} \neq \mathbf{0}$}}                                                                  & \multirow{2}{*}{$\|\Delta\hat{\mu}\|>0$} \\ \cline{3-3}       &                                                                                                      & $\hat{\mu}\lambda^{n} = \|\lambda^{t}\|$                                                                                & \multicolumn{1}{c|}{}                                                                                            &                                    & \multicolumn{1}{c|}{}                                                                                                    &                                    \\ \hline
\end{tabular}%
}
\label{table:CompareCase}
\end{table}

\subsection{Gradients of Contact Impulse}
In this section, we introduce the concept of the gradient of contact impulse with respect to the coefficient of friction, as described in the previous work~\cite{werling2021fast}.  Since the previous study briefly covered the gradient for the sliding state, we adopt more extended three-dimensional descriptions from~\cite{kim2023contactimplicit}. 

Given our framework's focus on estimating the friction coefficient, we consider only states where contact is detected by the state estimator~\cite{Joonha2023TRO}, excluding opening contacts. Whether the contact impulse from contact dynamics~\cite{raisim} touches the friction cone determines if the contact is sliding $\mathbf{s}$ or clamping $\mathbf{c}$. The contact Jacobian can be divided into $\mathbf{J}_\mathbf{c}$ and $\mathbf{J}_\mathbf{s}$, and the contact impulse into $\bm{\lambda}_\mathbf{c}$ and $\bm{\lambda}_\mathbf{s}$, depending on whether each contact index involves clamping or sliding~\cite{werling2021fast}.  For example, given $\mathbf{c}=\{1,3\}$, the corresponding contact Jacobian becomes $\mathbf{J}_{\mathbf{c}}=\left[\mathbf{J}_1^T,~\mathbf{J}_3^T\right]^T$ and corresponding contact velocity becomes $\mathbf{v}_{\mathbf{c},i+1}=\left[{\mathbf{v}_{1,i+1}}^T,~{\mathbf{v}_{3,i+1}}^T\right]^T$. The contact velocity~\eqref{eq:contactvel} can be expressed with the subscripts:
 \begin{align}
\label{eq:next_state_clamping}
\textbf{v}_{k,i+1}\nonumber&=\mathbf{J}_{k}\mathbf{M}^{-1}((-\mathbf{h}+\mathbf{B} \boldsymbol{\tau}_i)\Delta t+\mathbf{M}\boldsymbol{\upsilon}_i+{\mathbf{J}_\mathbf{c}}^T \bm{\lambda}_{\mathbf{c}}+{\mathbf{J}_\mathbf{s}}^T \bm{\lambda}_{\mathbf{s}}).
\end{align}

In the sliding state for $k\in\mathbf{s}$, the contact impulse is attached to the friction cone defined by the friction coefficient $\mu$:
\begin{equation}
\label{eq:slip_contact_impulse}
\bm{\lambda}_{k} =  \mathbf{E}_{k}{\lambda}^n_{k},
\end{equation}
where $\mathbf{E}_k = [-\mu\cos(\theta_k),  -\mu\sin(\theta_k), 1]^T$ and $\theta_{k}$ is the direction of the tangential contact velocity at $k$th contact. 

Considering the constraints in Fig.~\ref{figure:contact_dynamics}, contact velocities at clamping, $\textbf{v}_{\bm{c},i+1}$, and normal contact velocity at sliding $\textbf{v}^{n}_{\bm{s},i+1}$ are zero. By integrating these conditions with~\eqref{eq:contactvel}, the stacked contact impulse $\bm{\lambda}^{\mathrm{contact}}$ can be denoted as follows:
\begin{align}
    \label{eq:vcc_zero}
    \textbf{0} &= \mathbf{A}\bm{\lambda}^{\mathrm{contact}} + \mathbf{b},
\end{align}
where
\begin{align}
\bm{\lambda}&^{\mathrm{contact}}=\begin{bmatrix}
{\bm{\lambda}_\mathbf{c}}^T {\bm{\lambda}^n_\mathbf{s}}^T
\end{bmatrix}^T,
\nonumber\\
    \mathbf{A}&=\begin{bmatrix}
    \mathbf{J}_{\mathbf{c}}\nonumber\\
    \mathbf{J}^n_{\mathbf{s}}
    \end{bmatrix}
    \mathbf{M}^{-1}
    \begin{bmatrix}
    \mathbf{J}_{\mathbf{c}}\\
    \mathbf{E}^{T}_{\mathbf{s}}    \mathbf{J}_{\mathbf{s}}
    \end{bmatrix}^T,\nonumber\\
    \mathbf{b}&=\begin{bmatrix}
    \mathbf{J}_{\mathbf{c}}\\\mathbf{J}^n_{\mathbf{s}}
    \end{bmatrix}\mathbf{M}^{-1}\left((-\mathbf{h}+\mathbf{B} \boldsymbol{\tau}_{i})\Delta{t} + \mathbf{M}\boldsymbol{\upsilon}_i\right),\nonumber
\end{align}
and $\mathbf{E}_{\mathbf{s}}$ is a block diagonal matrix with top-left entry $\mathbf{E}_{s_1}$ and bottom-right entry $\mathbf{E}_{s_n}$. $s_1$ and $s_n$ are the first and last elements of set $\mathbf{s}$, respectively.
Then, the gradient of $\bm{\lambda}^{\mathrm{{contact}}}$ with respect to friction coefficient $\mu$ can be obtained:
\begin{align}
\label{non_smoothed_gradient1}
\frac{\partial \bm{\lambda}^{\mathrm{contact}}}{\partial \mu}= \mathbf{A}^{-1} \frac{\partial \mathbf{A}}{\partial \mu}\mathbf{A}^{-1}\mathbf{b}-\mathbf{A}^{-1}\frac{\partial \mathbf{b}}{\partial \mu}.
\end{align}

Note that the contact impulse in \eqref{non_smoothed_gradient1} depends on the coefficient of friction only when the contact state is sliding. When the sliding condition~\eqref{eq:slip_contact_impulse} is satisfied, the gradient of the contact impulse in the tangential direction can be expressed as follows: 
\begin{align}
\label{non_smoothed_gradient_sliding}
\frac{\partial \bm{\lambda}^{t}_\mathbf{s}}{\partial \mu} = \frac{\partial}{\partial \mu}(\mathbf{E}_{\mathbf{s}}\bm{\lambda}^n_{\mathbf{s}}).
\end{align}


\section{Method} \label{Sec:Method}

\begin{figure*}
    \centering
    \includegraphics[width=2.0\columnwidth]{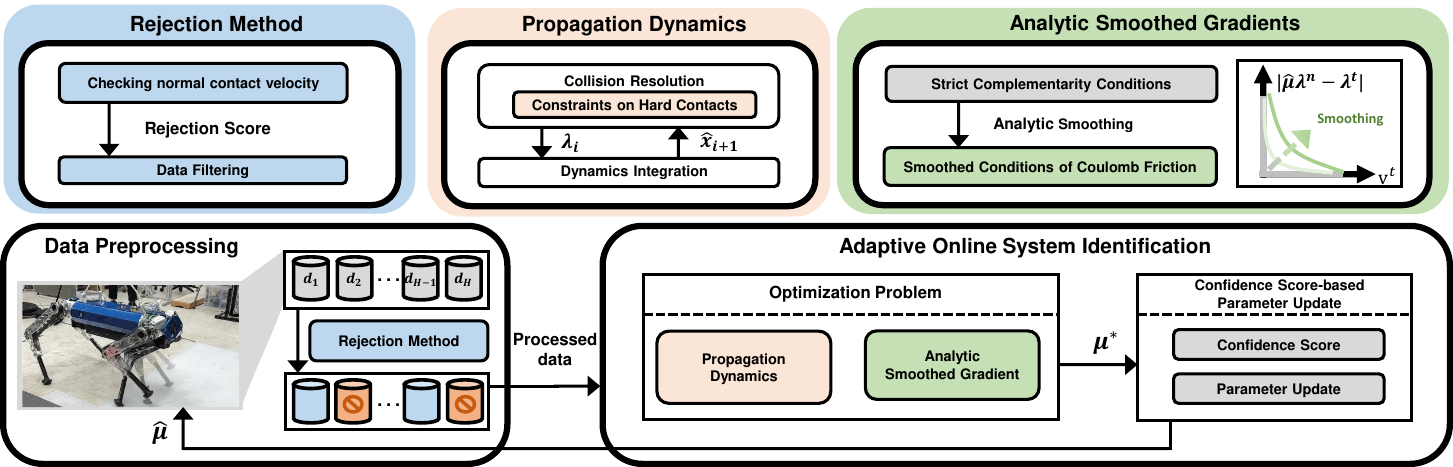}
    \caption{An overall proposed framework of the online friction coefficient identification for legged robots. Based on adaptive online system identification~\cite{chen2022real} using confidence score-based update, this work proposes analytic smoothed gradients with respect to friction coefficient and employs the rejection method. The rejection method calculates the rejection score based on the contact velocity in the normal direction and excludes the states where the rejection score exceeds a certain threshold. The processed data is utilized in the optimization problem, employing a hard contact model within the propagations of dynamics. When computing the gradient in the optimization problem, we specifically utilize the smoothed gradient of contact impulse with respect to the friction coefficient.}
    \label{figure:whole}
\end{figure*}

This section introduces our proposed method for analytic smoothed gradients of contact impulses with respect to the friction coefficient. Unlike previous works~\cite{kim2022contact,kim2023contactimplicit} that focused on smoothing the contact constraints in the normal direction, our approach applies smoothing to contact constraints in the tangential direction, as shown in the smoothed conditions of Fig.~\ref{figure:contact_dynamics}. We then detail our optimization formulation for friction coefficient identification and the rejection method with confidence score-based updates employed in our proposed framework. The overall framework is illustrated in Fig.~\ref{figure:whole}.

\subsection{Smoothed Contact Gradient}
Owing to the nature of the nonsmoothed constraints in rigid body contact dynamics, the analytic gradient of contact impulse with respect to the friction coefficient, in~\eqref{non_smoothed_gradient1} and~\eqref{non_smoothed_gradient_sliding}, can be non-informative. The lack of informative gradient often impedes the gradient-based strategies, posing a critical challenge for optimization~\cite{le2024leveraging}. Table~\ref{table:CompareCase} details how these non-informative gradients can hinder parameter updates of \eqref{osi_opt_basic}. When the estimated friction coefficient, $\hat{\mu}$, is higher than the actual one, $\mu_\mathrm{true}$, the contact in the dynamics model can be stuck at clamping, even if the robot slips. In this case, the contact impulse obtained from dynamics is not attached to the friction cone, as clamping conditions in Fig.~\ref{figure:contact_dynamics}, and the contact dynamics becomes independent of the friction coefficient~\cite{raisim}, leading to non-informative gradients. The non-informative gradient can prevent the optimized parameter in~\eqref{osi_opt_basic}, updated through the product of residuals and derivatives, from being updated to better local optima, even if the robot slips. As a result, the nonsmooth dynamics and its gradients can interrupt friction coefficient identification.

To tackle the lack of informative gradient, we propose analytic gradients with respect to the friction coefficient, derived by smoothing the complementarity constraint between contact velocity and Coulomb friction cone constraint in Fig.~\ref{figure:contact_dynamics}.

The complementarity constraint between contact velocity and Coulomb's friction cone constraint at $k$-th contact point, which is not a separating state, can be expressed as follows:
\begin{align}
    0\leq\label{non_smoothing_tangential_cond_complementarity}
    &\|\textbf{v}^{t}_{k}\| \perp{(\mu^2(\lambda^{n}_{k})^2-\|\boldsymbol{\lambda}^{t}_{k}\|^2)}\geq0.
\end{align}

We propose the smoothed constraint of~\eqref{non_smoothing_tangential_cond_complementarity}, expressed with the smoothing parameter $\rho_\mathrm{t}>0$ as:
\begin{align}
    \label{smoothing_tangential_cond}
    &\|\textbf{v}^{t}_{k}\| \cdot (\mu^2(\lambda^{n}_{k})^2-\|\boldsymbol{\lambda}^{t}_{k}\|^2)=\rho_\mathrm{t}.
\end{align}
We define the vector $\mathbf{\Theta}_{k}=[\cos(\theta_{k}),\sin(\theta_{k})]^{T}$ to represent the direction of contact velocity at the $k$th contact index. Separating the tangential contact velocity vector, $\textbf{v}^{t}_{k}\in\mathbb{R}^2$,  into the direction and magnitude gives the following expression:
\begin{align}
    &\textbf{v}^{t}_{k} = 
     \mathbf{\Theta}_{k}\|\textbf{v}^{t}_{k}\|=
     \mathbf{\Theta}_{k}\frac{\rho_\mathrm{t}}{(\mu^2(\lambda^{n}_{k})^2-\|\boldsymbol{\lambda}^{t}_{k}\|^2)}.
\end{align}
From the linear relation of contact velocities and impulses~\eqref{eq:contactvel},
\begin{align}
    \label{eq:tangential_implicit_differentiation}
     \mathbf{\Theta}_{k}\frac{\rho_\mathrm{t}}{(\mu^2(\lambda^{n}_{k})^2-\|\boldsymbol{\lambda}^{t}_{k}\|^2)} = A^t_{k} \boldsymbol{\lambda} + b^t_k.
\end{align}

Differentiating \eqref{eq:tangential_implicit_differentiation} with respect to the parameter $\mu$ gives the following relation of $\frac{\partial{\lambda}}{\partial \mu}$ and ${\lambda}$:
\begin{align}
    \label{tangential_smoothed_contact_gradient_equation}
   A^t_{k} \frac{\partial \boldsymbol{\lambda}}{\partial \mu}+\mathbf{\Theta}_{k} \hat{\rho}_{k} 
    \begin{bmatrix} -2{\boldsymbol{\lambda}^{t}_{k}}^T\ 2\mu^2\lambda^{n}_{k} \end{bmatrix} \frac{\partial \boldsymbol{\lambda}_k}{\partial \mu}\nonumber\\
    = - \frac{\partial A^t_{k}}{\partial \mu} \boldsymbol{\lambda} - \frac{\partial b^t_k}{\partial \mu}-\mathbf{\Theta}_{k}\hat{\rho}_{k} 2\mu(\lambda^{n}_{k})^2,
\end{align}
where $\hat{\rho}_{k}=\frac{\rho_\mathrm{t}}{(\mu^2(\lambda^{n}_{k})^2-\|\boldsymbol{\lambda}^{t}_{k}\|^2)^2} \in \mathbb{R}$.

Stacking the gradient equations for all contact points, we get the following linear system for the contact impulse:
\begin{align}
    \label{eq:smoothing_gradient}
    & (A + \mathbf{\Gamma}(\rho_\mathrm{t}) )\frac{\partial\bm{\lambda}}{\partial\mu} 
    =
    \begin{bmatrix} 
    \cdots \\
    - \frac{\partial A^t_{k}}{\partial \mu} \boldsymbol{\lambda} - \frac{\partial b^t_k}{\partial \mu}-\mathbf{\Theta}_{k} \hat{\rho}_{k} 2\mu(\lambda^{n}_{k})^2\\
    - \frac{\partial A^n_{k}}{\partial \mu} \boldsymbol{\lambda} - \frac{\partial b^n_k}{\partial \mu}\\
    \cdots 
    \end{bmatrix},
\end{align}
where $\mathbf{\Gamma}(\rho_\mathrm{t})$ is the block diagonal matrix of 
$ \mathbf{\Gamma}_{k}(\rho_\mathrm{t})=
\begin{bmatrix}
    \mathbf{\Theta}_{k} \hat{\rho}_{k} 
    \begin{bmatrix} -2{\boldsymbol{\lambda}^{t}_{k}}^T\ 2\mu^2\lambda^{n}_{k} \end{bmatrix} \\
    \mathbf{0}_{1\times3}
\end{bmatrix}\in\mathbb{R}^{3\times3}$.\\



Finally, we simplify the gradient equation as follows:
\begin{align}
    &\frac{\partial \bm{\lambda}}{\partial \mu}=-[\mathbf{A}+\mathbf{\Gamma}(\rho_\mathrm{t})]^{-1}(\frac{\partial \mathbf{A}}{\partial \mu} \bm{\lambda}+\frac{\partial \mathbf{b}}{\partial \mu} + \mathbf{\gamma}(\rho_\mathrm{t})), \label{eq:smoothing_gradient2}
\end{align}
where $\mathbf{\gamma}(\rho_\mathrm{t})$ is stacked by $\mathbf{\gamma}_{k}(\rho_\mathrm{t})=
\begin{bmatrix} 
\mathbf{\Theta}_{k}\hat{\rho}_{k}2\mu(\lambda^{n}_{k})^2\\
    0
\end{bmatrix}\in\mathbb{R}^{3}$.

Compared to the nonsmooth gradient as~\eqref{non_smoothed_gradient1} and~\eqref{non_smoothed_gradient_sliding}, $\mathbf{\Gamma}(\rho_\mathrm{t})$ and $\mathbf{\gamma}(\rho_\mathrm{t})$ are additional terms for the smoothed constraints.

\subsection{Optimization Problem for Friction Coefficient Identification}

\subsubsection{Problem Definition}
Consider the friction coefficient $\mathbf{\mu}$, and a system with discrete time dynamic model:
\begin{equation}
\label{osi_dyn}
     \hat{\mathbf{x}}_{i+1} = {f}(\mathbf{x}_i,\boldsymbol{\tau}_i,\mathbf{\bm{\lambda}}(\mathbf{x}_i,\boldsymbol{\tau}_i,\mu)),
 \end{equation}
 where ${f}$ is the dynamics function, based on~\cite{raisim}.
 
 In this study, based on~\eqref{osi_opt_basic}, we address an optimization problem to find the optimal parameters $\mu^{*}$ as follows:
\begin{equation}
    \begin{aligned}
\label{osi_opt}
     \mu^{*} = \arg\min_{\mu} &\sum_{i=1}^{H-1} \left\| {f}(\mathbf{x}_i,\boldsymbol{\tau}_i,\mathbf{\bm{\lambda}}(\mathbf{x}_i,\boldsymbol{\tau}_i,\mu)) - \mathbf{x}_{i+1} \right\|_{\Sigma}\\\quad  
      &s.t. \quad \mu_\mathrm{{min}} \leq \mu \leq \mu_\mathrm{max},
         \end{aligned}
 \end{equation}
 where the weighting matrix is denoted by $\Sigma$. $\Sigma$ is defined as $\mathrm{\diag}(\sigma_{q_\mathrm{base}},\sigma_{q_\mathrm{jnt}},\sigma_{\dot{q}_\mathrm{base}},\sigma_{\dot{q}_\mathrm{jnt}})$, which represents a diagonal matrix of weight factors for the base pose, joint angles, base velocity, and joint velocities, respectively. If the norm of tangential contact velocity exceeds 0.4~\si{\meter/\second}, the parameters $\sigma_{q_\mathrm{jnt}}$ and $\sigma_{\dot{q}_\mathrm{jnt}}$ are scaled by $\sigma_\mathrm{slip}$. $\mu_{\mathrm{min}}$ and $\mu_{\mathrm{min}}$ are the lower and upper bounds for the estimated coefficient of friction.
  
In this work, the Sequential Quadratic Programming Gauss-Newton method with a Hessian approximation is used to address the nonlinear least-squares problem of~\eqref{osi_opt}. 



\subsection{Confidence Score-Based Parameter Update}
\label{sec:conf}
This study employs a confidence score-based update method proposed in~\cite{chen2022real}. This method enables the improvement of online system identification by assessing parametrically exciting observations when identifying the parameter. For instance, when a legged robot does not slip, the state history provides limited information about the friction coefficient~\cite{focchi2018slip}. In contrast, data from a sliding system are more informative for identifying this coefficient.
In this way, friction coefficient identification can be improved by leveraging the confidence score, which enables distinguishing between the non-informative and informative observations~\cite{chen2022real}. In the previous work~\cite{chen2022realarxiv}, a confidence score for the friction coefficient was proposed. The score increases when the tangential contact velocity is nonzero. This includes cases where legged robots have a high contact velocity following contact initiation, which can undesirably increase the score on nonslippery terrain.

In this work, we employ both a confidence score and rejection method, directly based on the contact velocity. In this section, we first introduce the confidence score. The rejection method will be described in Sec.~\ref{sec:rej}.

As in~\cite{chen2022realarxiv}, we define the confidence score $\eta$, based on tangential contact velocity, for friction coefficient identifications:
\begin{align}
\label{eq:confidence_score}
\eta &= 1-\exp(-\alpha_\mathrm{conf}\mathbf{v}^{t}_\text{mean}),
\end{align}
where $\alpha_\mathrm{conf}$ represents a positive constant, and $\mathbf{v}^{t}_\text{mean}$ denotes the average norm of nonzero tangential contact velocity in the data buffer, with rejected data excluded.

As described in~\cite{chen2022real}, estimated parameters are not updated until the score exceeds a threshold, $\gamma_\mathrm{conf}$. If the difference between the current estimate $\hat{\mu}$ and the optimum $\mu^{*}$ from~\eqref{osi_opt} exceeds a specific threshold $\epsilon$, the estimate is directly updated to this optimum. Otherwise, the update is performed through a weighted sum using the previous confidence score.


\subsection{Data Preprocessing}
\subsubsection{{Data Buffer}}
\label{eq:databuffer}
For real-time parameter identification, data including states are collected at each time step. Let $\mathbf{d}_{i}$ be the ${i}$-th data in the buffer, collected at each time step $\Delta{t}_\mathrm{buffer}$. We define $\mathbf{d}_{i}$ as follows:
\begin{align}
\mathbf{d}_{i} = \begin{bmatrix}\mathbf{R}_{i}, \mathbf{p}_{i}, \boldsymbol{\omega}_{i}, \mathbf{\dot{p}}_{i}, \mathbf{q}_{\mathrm{jnt},i}, \mathbf{\dot{q}}_{\mathrm{jnt},i}, \boldsymbol{\tau}_{i}, \mathbf{c}_{i}, \mathbf{v}_{i}\end{bmatrix},
\end{align}
where $\mathbf{R}_{i}\in\text{SO}(3)$ represents the rotation matrix for the robot's base, $\mathbf{p}_{i}$ is the position of the robot, and $\boldsymbol{\omega}_{i}$ is the angular velocity of the base, $\mathbf{\dot{p}}_{i}$ is the linear velocity of the robot, $\mathbf{q}_{\mathrm{jnt},i}$ is the joint angle, $\mathbf{\dot{q}}_{\mathrm{jnt},i}$ is the joint angular velocity, $\boldsymbol{\tau}_{i}$ represents the joint torque, and $\mathbf{v}_{i}$ is the estimated foot velocity. $\mathbf{c}_{i}$ represents the estimated contact state. 
The data buffer is comprised of the set of $\mathbf{d}_{i}$, where $i=1,\cdots,H$. When the buffer reaches its capacity, the oldest collected data will be removed to accommodate the saving of new data.

\subsubsection{Data Rejection Method}
\label{sec:rej}
Raw data from legged robots can often include states with high normal and tangential contact velocities following contact initiation. Such velocities can lead to undesirable increases in confidence scores for the friction coefficient as the score rises with the tangential contact velocity. For example, when the robot navigates on nonslippery terrain, the confidence score can increase due to high contact velocity following contact initiation. Consequently, utilizing the data for friction coefficient identification can lead to undesirably high confidence scores and inconsistent parameter updates, especially on nonslippery terrain.

To address these issues, this work employs a data rejection method.
We define a rejection score to evaluate the extent of  contact velocity in the normal direction for $i=1,\cdots,H$:
\begin{align}
\label{eq:rejection_score1}
r_{k,1,i} &= ( 1 - {\text{c}}_{k,i}\exp(-\alpha_\mathrm{rej}\text{v}^{n}_{k,i}) ), \\
\label{eq:rejection_score2}
r_{k,2,i} &= |r_{k,1,i} - r_{k,1,i-1}|,\\
\label{eq:rejection_score}
r_{k,i} &= \max(r_{k,1,i},r_{k,2,i}),
\end{align}
where $k$ is the index of contact points, $\alpha_\mathrm{rej}$ and $\beta_\mathrm{rej}$ are positive constants.
The first rejection score~\eqref{eq:rejection_score1} monitors the $k$-th normal contact velocity at $i$-th data in the data buffer. The second rejection score~\eqref{eq:rejection_score2} assesses the changes of the normal contact velocity, described using two consecutive indices. $r_{k,1,0}$ and $r_{k,2,0}$ are set as zero. If $k$-th contact's rejection score of \eqref{eq:rejection_score} exceeds a threshold $\gamma_\mathrm{rej}$, the $k$-th contact will be excluded from friction coefficient identification.


\begin{figure*}
    \centering
\includegraphics[width=\textwidth]{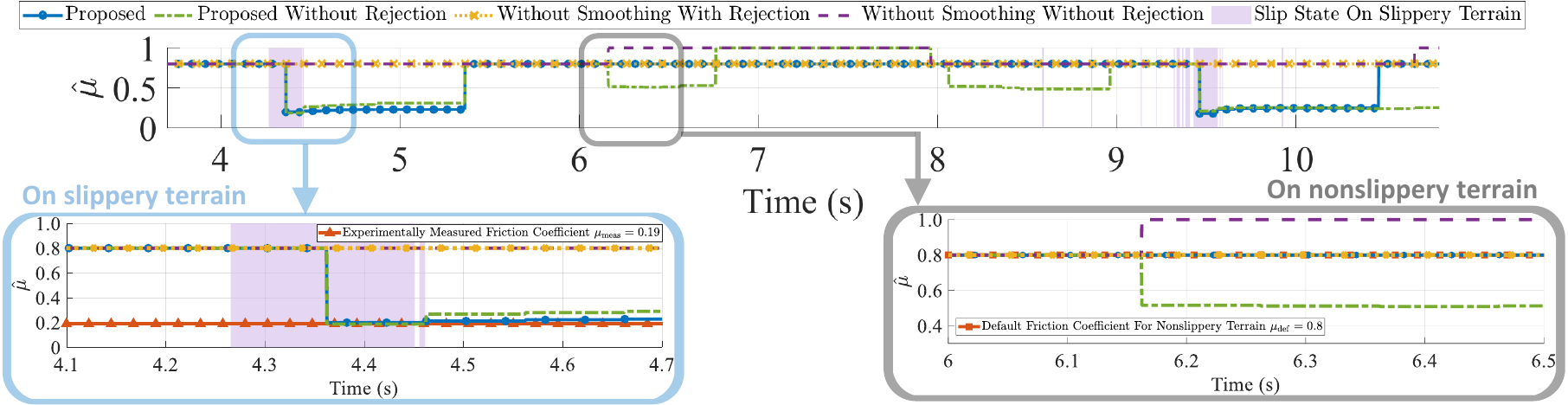}
    \caption{The experimental results of friction coefficient identification show the effects of proposed smoothed gradients and rejection methods. Without smoothed gradients, non-informative gradients can impede friction coefficient identification. The rejection method allows for consistent friction coefficient identification, especially when the legged robot traverses nonslippery terrain. The purple area represents the slip states on the slippery terrains where the norm of tangential estimated foot velocity from the state estimator~\cite{Joonha2023TRO} exceeds 0.4~\si{\meter/\second}.}
    \label{fig:online_estimation}
\end{figure*}


\begin{figure}
   \centering
\includegraphics[width=1.0\columnwidth]{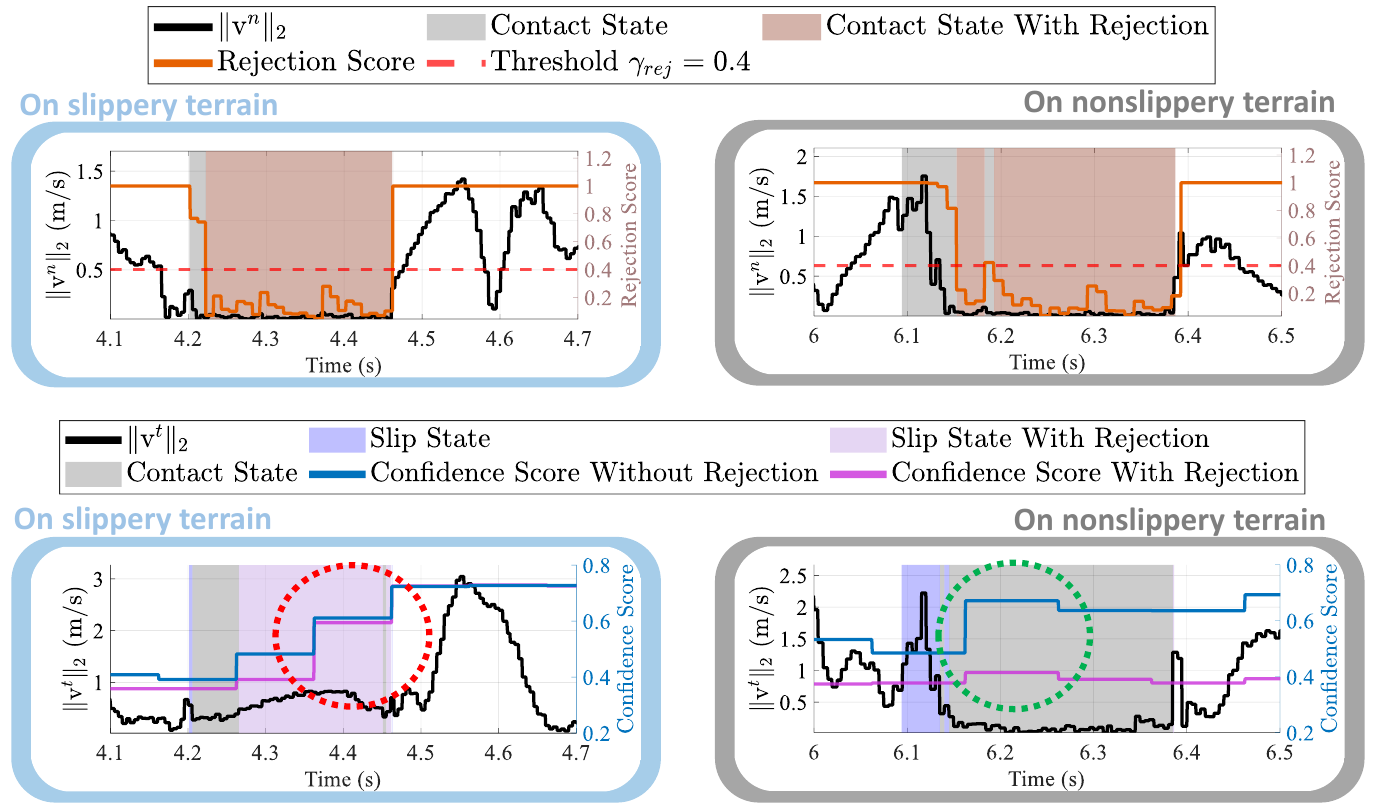}
   \caption{As depicted by a green dotted circle, the rejection method can effectively prevent undesired increases in the confidence score, especially on nonslippery terrain. Conversely, described by a red dotted circle, rejection scores do not significantly impede the increases in confidence scores when the robot slips on the slippery terrain.}
   \label{fig:contact_vel}
\end{figure}

\begin{figure}
    \centering
    \includegraphics[width=1.0\columnwidth]{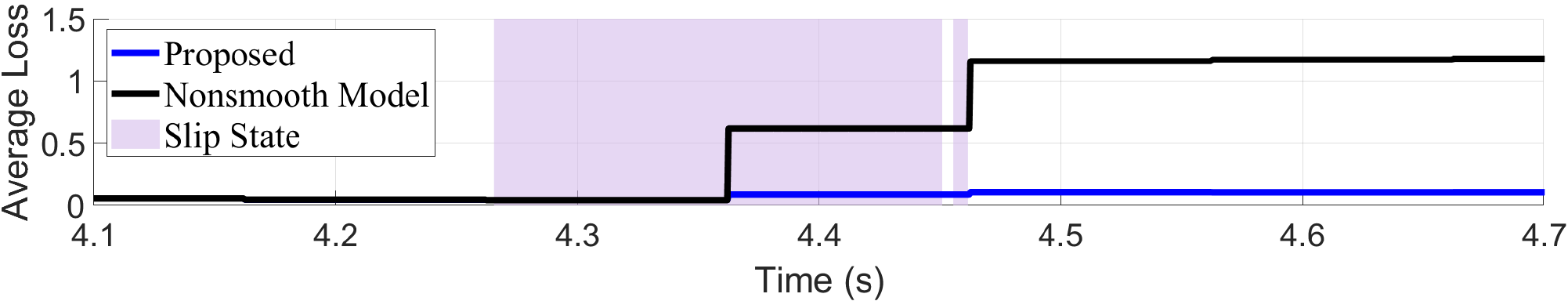}
    \caption{Comparison of the average loss between the method using the nonsmooth model and the proposed method. The proposed method achieves a lower average loss than the method using the nonsmooth model.}
    \label{figure:loss_compare}
\end{figure}

\section{Experimental Results} \label{Sec:Experiment}

\begin{table}[]
\caption{Parameters Used in Experiments}
\centering
\resizebox{\columnwidth}{!}{%
\begin{tabular}{|c|c|c|c|c|c|c|c|}
\hline
\hline
\textbf{Parameter} & \text{$\alpha_\mathrm{rej}$} & \text{$\gamma_\mathrm{rej}$} & \text{$\Delta{t}_\mathrm{buffer}$} & \text{$\Delta{t}_\mathrm{bound}$} & \text{$\sigma_\mathrm{slip}$} & \text{$\sigma_{q_\mathrm{base}}$} & \text{$\sigma_{q_\mathrm{jnt}}$} \\ \hline
\textbf{Value}     &   5.0    &     0.4    &             0.01~\si{\second}       &      0.1~\si{\second}    & 30 & 1e-4 & 20      \\
\hline \hline
\textbf{Parameter} & \text{$\alpha_\mathrm{conf}$} & \text{$\gamma_\mathrm{conf}$} & \text{$\epsilon$} & \text{$H$} & \text{$\rho_\mathrm{t}$} & \text{$\sigma_{\dot{q}_\mathrm{base}}$} & \text{$\sigma_{\dot{q}_\mathrm{jnt}}$}\\ \hline
\textbf{Value}     &    3.0       &       0.58     &     0.1      &      50       &      0.05      & 1e-4 & 1    \\
\hline \hline
\end{tabular}%
}
    \label{table:parameters}
\end{table}
This section introduces the experimental results of the proposed friction coefficient identification framework using the quadrupedal robot, KAIST HOUND~\cite{shin2022hound}. Additionally, we explain the effects of proposed analytic smoothed gradients of contact impulses with respect to the friction coefficient and the proposed rejection method. 
\subsection{Experimental Setup}

The proposed framework is based on the confidence score-based online system identification framework~\cite{chen2022real} and employs the proposed smoothed gradient and rejection method. To calculate the proposed smoothed gradients, we empirically set the smoothing parameter, $\rho_\mathrm{t}$, as 0.05. As~\cite{kim2023contactimplicit}, if the smoothing parameter is either too large or too small, the smoothing method may not be effective in achieving better local optima compared to the nonsmooth method. In implementing the proposed framework, we set $\mu_\mathrm{min}$ and $\mu_\mathrm{max}$ as 0.01 and 1.0, respectively.

Since estimating a high friction coefficient through contact dynamics is challenging in the absence of foot slippage~\cite{focchi2018slip,jenelten2019legged}, this work employs the reset method for the friction coefficient as~\cite{ jenelten2019legged}. This method resets the estimated friction coefficient to the default value $\mu_\mathrm{def}$ of 0.8 when stable contacts are maintained for 0.5~\si{\second}. In this work, the estimated friction coefficient is restored to the default value if the confidence score $\eta$ does not exceed the threshold $\gamma_\mathrm{conf}$ for 0.5~\si{\second}, indicating that the robot does not have a high tangential contact velocity for this duration.

The experiments are conducted in two different terrains: a nonslippery terrain and a slippery terrain. The slippery terrain is made of acrylic flat boards with boric acid powder. The robot initially starts on the nonslippery terrain where the experimentally measured friction coefficient is 1.0, then moves to the slippery terrain where the experimentally measured friction coefficient is 0.19. Subsequently, the robot moves between slippery and nonslippery terrains alternately. The measured friction coefficient on slippery terrain was obtained by measuring the horizontal force with a spring scale when the standing robot began to slip, considering its weight~\cite{shin2022hound}.

We solved the contact dynamics only for the states where contact is detected by the state estimator and implemented RaiSim’s algorithm~\cite{raisim} for this purpose.

For state estimation of the legged robot, we employ the method proposed by \cite{Joonha2023TRO}, which operates at 200 Hz within our framework. The contact velocity, contact states, and slip states are estimated in the state estimator. We determine the slip states when the norm of tangential contact velocity, estimated by the state estimator exceeds 0.4~\si{\meter/\second}. As a robot's controller, a nonlinear model predictive controller in \cite{hong202realtime} is utilized with functioning at 80 Hz. The boundary of computation time ${\Delta{t}_\mathrm{bound}}$ for the proposed framework is set at 10 Hz. The detailed parameters for the proposed framework are given in Table.~\ref{table:parameters}. A single onboard computer with an Intel(R) Core(TM) i7-11700T CPU, capable of reaching up to 1.6 GHz, is utilized to implement the proposed framework.

\subsection{Estimation Results}

\begin{figure}
    \centering
    \includegraphics[width=1.0\columnwidth]{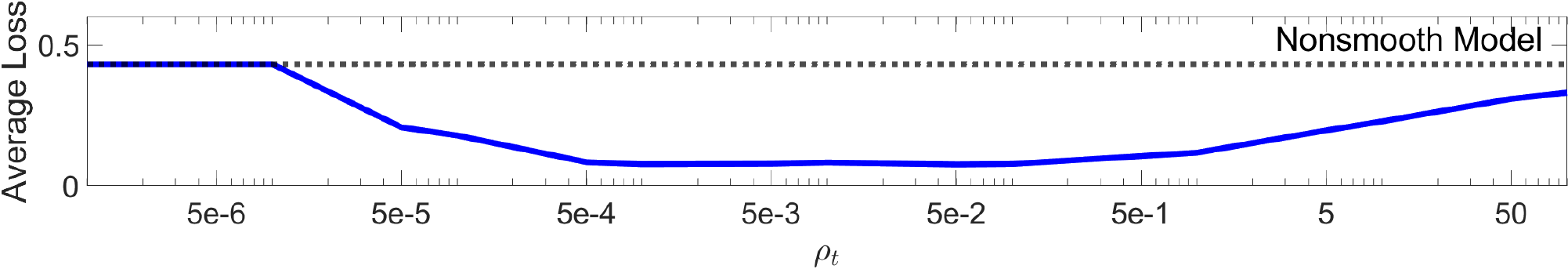}
    \caption{Result of the average loss for the experiment according to the smoothing parameter $\rho_\mathrm{t}$.}
    \label{figure:smoothing_compare}
\end{figure}

To validate the proposed methods, we compared the results of friction coefficient identification with and without the proposed gradient and rejection method, as shown in Fig.~\ref{fig:online_estimation}. In the experiment, we set the default estimated friction coefficient to 0.8 to illustrate a scenario where the robot, assuming a high friction coefficient for non-slippery terrain, slips on slippery surfaces. Note that the parameter update is conducted when the confidence score exceeds the threshold $\gamma_\mathrm{conf}$~\cite{chen2022real}.


In the left bottom figure of Fig.~\ref{fig:online_estimation}, we observed that using nonsmooth gradients can impede friction coefficient identification, even if the robot slips on slippery terrains. In contrast, employing the proposed smoothing method allows for fast and consistent identification.

Moreover, the right bottom figure in Fig.~\ref{fig:online_estimation} shows that the estimated friction coefficient becomes more consistent, especially on nonslippery terrain, when the confidence score-based update is used with the rejection method compared to without it. Using both methods, the estimated friction coefficient on nonslippery terrain can be maintained close to the default value for such terrain, without undesired updates.



The detailed effects of the proposed gradients and rejection methods will be discussed below.

\subsection{The Effects of Analytic Smoothed Contact Gradients}
\begin{figure}
    \centering
    \subfloat[]{
        \includegraphics[width=0.45\columnwidth]{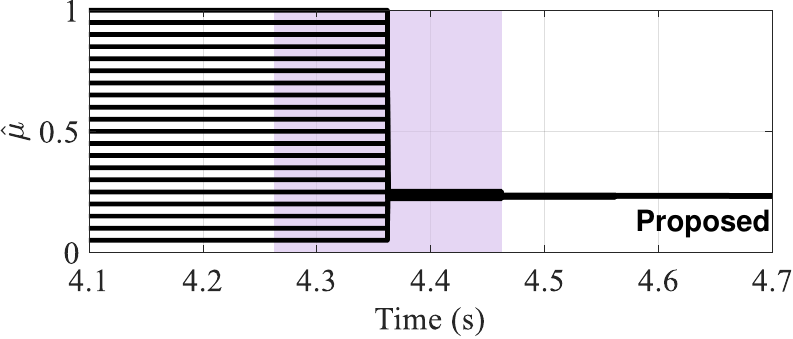}
        \label{figure:different_initial_condition_prop}
        }
        \hfill
    \subfloat[]{
        \includegraphics[width=0.45\columnwidth]{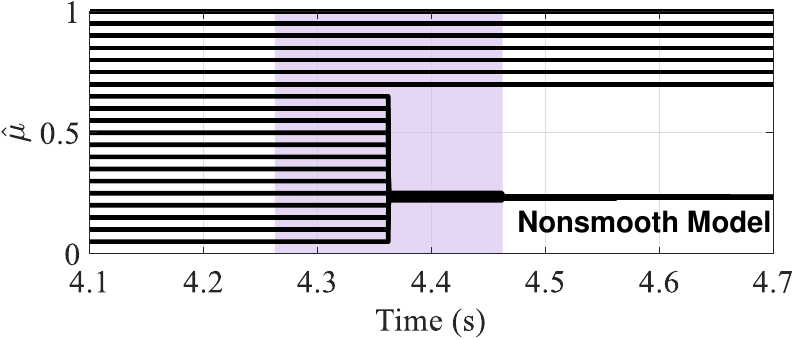}
        \label{figure:different_initial_condition_nonsm}
        }
    \caption{Comparison of friction coefficient identification under various initial estimates. The purple area represents the slip states on slippery terrains where the norm of tangential contact velocity exceeds 0.4~\si{\meter/\second}. \protect\subref{figure:different_initial_condition_prop} The proposed smoothing is applied. \protect\subref{figure:different_initial_condition_nonsm} The gradient from the nonsmooth model is used~\cite{chen2022real}.}
    \label{figure:different_initial_condition}
\end{figure}

In this session, we will examine the advantages of the proposed smoothing method in friction coefficient identification. As shown in Fig.~\ref{fig:online_estimation}, when the robot slips on slippery terrain, the proposed smoothing method enables parameter updates towards a low friction coefficient, in contrast to the case of the nonsmooth model. For the slipping case, we compare the average loss of the nonsmooth model with that of the smoothing method in Fig.~\ref{figure:loss_compare}. In the figure, we observed that using the proposed smoothing method can lead to convergence at better local optima, achieving a lower loss value. Specifically, Fig.~\ref{figure:smoothing_compare} shows the average loss during the experiments shown in Fig.~\ref{fig:online_estimation} according to the smoothing parameter. We observed that when the smoothing parameter $\rho_\mathrm{t}$ is excessively increased or decreased, the effect for the convergence towards a lower loss may be reduced, as~\cite{kim2023contactimplicit}.

Moreover, we conducted friction coefficient identification with various initial conditions in 0.05 units from 0.05 to 1.0, as shown in Fig.~\ref{figure:different_initial_condition}.
In the experiment, we used the same experimental data as that for Fig.~\ref{fig:online_estimation}. We compared the performance of friction coefficient identification between the proposed model and the nonsmooth model on slippery terrain. As shown in Fig.~\ref{figure:different_initial_condition_nonsm}, 
when employing nonsmooth gradients, the lack of informative gradients can lead to the failure to identify the lower friction coefficient. We observed that the issue often occurs as the gap between the estimated friction coefficient and the actual one is large. In contrast, as Fig.~\ref{figure:different_initial_condition_prop}, our proposed smoothing method solves the failure issue of parameter identification, even under various initial conditions.

Considering the results, we observed that the proposed smoothing method provides advantages for friction coefficient identification under various initial conditions, even when a high initial friction coefficient leads to non-informative gradients. These advantages can be utilized in various model-based frameworks. For instance, model-based controllers for legged robots often employ a user-defined friction coefficient to compute control inputs based on the Coulomb friction cone constraint. The friction coefficient is typically determined by heuristic tuning for their tasks~\cite{jenelten2019legged,hong202realtime}. A high friction coefficient can be selected to optimize control inputs, leveraging more tangential ground reaction forces. However, using a high friction coefficient on slippery terrain may cause the robot to slip, as the control inputs are computed based on a high friction coefficient. Consequently, there is a need for real-time friction coefficient identification that performs fast and consistently on slippery terrain, even with a high initial friction coefficient. The proposed framework can identify the friction coefficient under various initials, handling non-informative gradients.

\subsection{Comparison with Randomized Smoothing Methods}
\begin{figure}
    \centering
    \includegraphics[width=1.0\columnwidth]{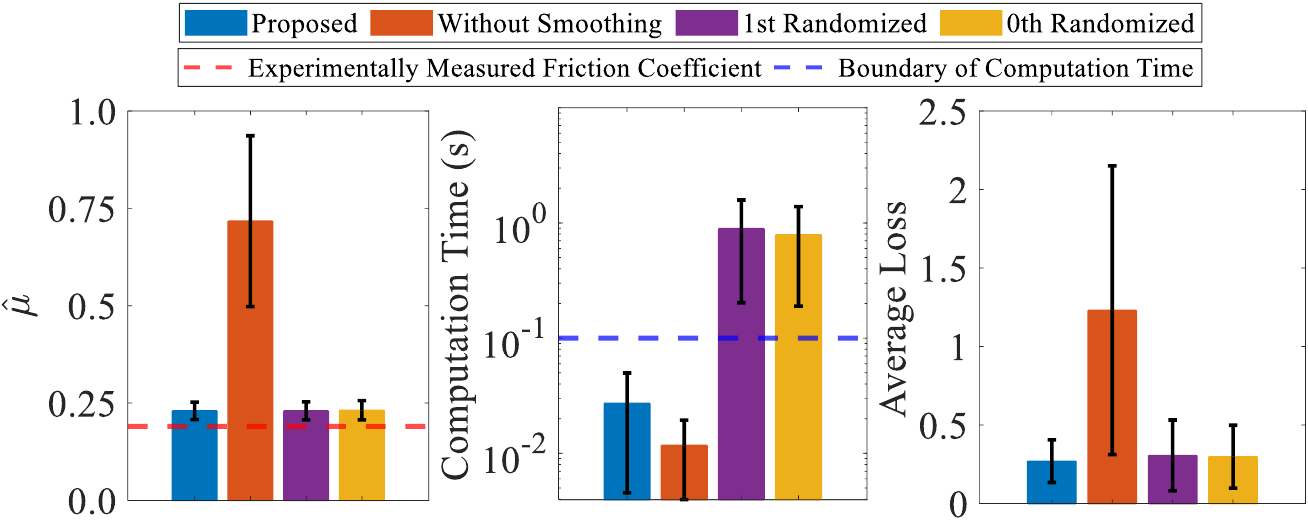}
    \caption{Compared to the baselines, the proposed methods can achieve fast and consistent friction coefficient identification in real-time.}
    \label{figure:compare_randomize}
\end{figure}

In this section, we compare the performance of friction coefficient identification using the proposed gradients with baseline methods. For the baselines, we adopt the online system identification using nonsmooth gradients~\cite{chen2022real} and using randomized smoothing methods~\cite{le2024leveraging, Pang2023TRO}: specifically first-order and zeroth-order randomized smoothing methods. The randomized smoothing methods utilize 50 samples with parallel computing to obtain stochastic gradients. We conducted seven experiments where the robot slipped on slippery terrains, with initial estimates of 0.8. 

The results are summarized in Fig.~\ref{figure:compare_randomize}, which presents histograms of the estimated friction coefficient, computation time for solving the optimization problem for~\eqref{osi_opt}, and average loss. We observe that the proposed smoothed gradient results in lower computation times than other randomized smoothing methods. As noted in~\cite{Pang2023TRO}, while randomized smoothing methods can address the lack of informative gradients, they require longer computation times due to sampling. Furthermore, it is observed that the mean and standard deviation of estimates without the smoothing method are higher than those using smoothing methods. This can be attributed to the lack of informative gradients, which causes the gradient-based optimization strategy to fail in friction coefficient identification.

\subsection{The Effects of Data Rejection Method}

In this section, we describe the benefits of data rejection methods by comparing updates based on confidence scores with and without rejection methods. In the right bottom figure of Fig.~\ref{fig:online_estimation}, friction coefficient identification with the rejection method is more consistent than without it, especially on nonslippery terrain.

As illustrated in the bottom right figure of Fig.~\ref{fig:contact_vel}, not using the rejection method can lead to an increased confidence score, even on nonslippery terrain. If the confidence score increases on nonslippery terrain, the parameter updates can be conducted using non-informative observations, leading to undesired and inconsistent friction coefficient identification.

However, with the proposed rejection method, the confidence score on nonslippery terrain does not increase as much as it does without the method, allowing for consistent performance in friction coefficient identification. Furthermore, the rejection method does not significantly impede increases in the confidence score when the robot slips. As shown in the bottom-left of Fig.~\ref{fig:contact_vel}, when the robot slips on slippery terrain, the confidence score with the rejection method is comparable to one without the method. The upper figures of Fig.~\ref{fig:contact_vel} show that the data with high contact velocity following contact initiations can be excluded from parameter identification.

\section{Discussion}\label{Sec:Discussion}
In this section, we explain the limited scope of nonsmooth dynamics derivatives from Section~\ref{Sec:Background} and outline future work. The nonsmooth gradients derived through time-stepping methods may be incorrect or non-informative~\cite{kong2024saltation,le2024leveraging,werling2021fast}. This work specifically addresses the issue of non-informative gradients by employing smoothing approaches. By comparison, to obtain correct gradients for nonsmooth discrete-time systems, discrete event timing variations should be accounted for using the Saltation matrix~\cite{kong2024saltation}. In future work, we plan to use the Saltation matrix to obtain correct gradients for dynamic parameter identification and compare the results with those of proposed smoothed gradients.

\section{Conclusion} \label{Sec:Conclusion}
We presented an online friction coefficient identification framework for legged robots on slippery terrains using the proposed analytic smoothed gradient of contact impulse with respect to the friction coefficient. The experimental results showed that the proposed smoothed gradient allows for overcoming the issue of non-informative gradients in friction coefficient identification. We observed that the framework using the proposed smoothed gradients shows less computation time in experiments than using randomized smoothing methods. Moreover, the rejection method improved consistency in friction estimation over existing system identification~\cite{chen2022real}. This framework could benefit model-based frameworks that require an online estimated friction coefficient for legged robots. 



\bibliographystyle{ieeetr}
\bibliography{IEEEabrv, reference}

\end{document}